\documentclass[10pt,twocolumn,letterpaper]{article}

\usepackage{cvpr}
\usepackage{times}
\usepackage{graphicx}
\usepackage{amsmath}
\usepackage{amssymb}
\usepackage{comment}
\usepackage{microtype}
\usepackage{multirow,tabularx}
\usepackage{tikz}
\usepackage{enumitem}
\usetikzlibrary{positioning,3d}
\usetikzlibrary{decorations.pathreplacing}

\cvprfinalcopy 

\def\cvprPaperID{****} 

\begin{document}

\title{Network Dissection: \\Quantifying Interpretability of Deep Visual Representations}

\author{
David Bau\footnotemark,~~Bolei Zhou\footnotemark[\value{footnote}],~~Aditya Khosla,~~Aude Oliva,~~and Antonio Torralba \\
CSAIL, MIT\\
\texttt{\{davidbau, bzhou, khosla, oliva, torralba\}@csail.mit.edu}
}

\maketitle
\global\csname @topnum\endcsname 0
\global\csname @botnum\endcsname 0
\begin{abstract}
\vspace{-1mm}
We propose a general framework called \textit{Network Dissection} for quantifying the interpretability of latent representations of CNNs by evaluating the alignment between individual hidden units and a set of semantic concepts. Given any CNN model, the proposed method draws on a broad data set of visual concepts to score the semantics of hidden units at each intermediate convolutional layer. The units with semantics are given labels across a range of objects, parts, scenes, textures, materials, and colors. We use the proposed method to test the hypothesis that interpretability of units is equivalent to random linear combinations of units, then we apply our method to compare the latent representations of various networks when trained to solve different supervised and self-supervised training tasks.  We further analyze the effect of training iterations, compare networks trained with different initializations, examine the impact of network depth and width, and measure the effect of dropout and batch normalization on the interpretability of deep visual representations. We demonstrate that the proposed method can shed light on characteristics of CNN models and training methods that go beyond measurements of their discriminative power.
\vspace{-5mm}
\end{abstract}

\section{Introduction}
\begin{figure}
\begin{center}
{\sffamily\scriptsize\def\arraystretch{1}
\begin{tabularx}{\columnwidth}{@{}X@{\hspace{.4em}}X@{\hspace{.4em}}X@{}}
lamps in places net%
& wheels in object net%
& people in video net \\
\parbox[c]{2.6cm}{\includegraphics[width=2.6cm,height=1.3cm,trim={0 0 5690px 0},clip]{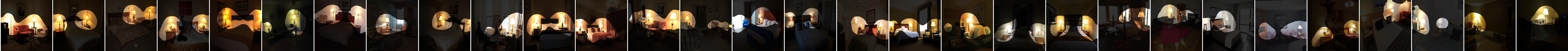}}
& \parbox[c]{2.6cm}{\includegraphics[width=2.6cm,height=1.3cm]{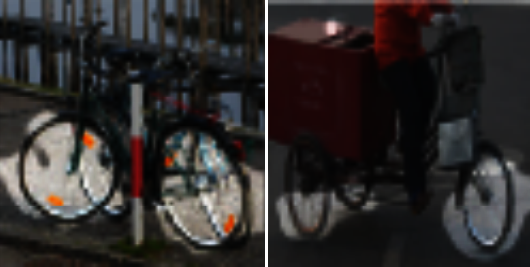}}
& \parbox[c]{2.6cm}{\includegraphics[width=2.6cm,height=1.3cm]{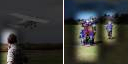}}
\\[9mm]
\end{tabularx}%
}

\end{center}
\vspace{-6mm}
\caption{Unit 13 in \cite{zhou2014object} (classifying places) detects table lamps. Unit 246 in \cite{gonzalez2016semantic} (classifying objects) detects bicycle wheels. A unit in \cite{vondrick2016generating} (self-supervised for generating videos) detects people.}
\label{cited-figure}
\vspace{-4mm}
\end{figure}

Observations of hidden units in large deep neural networks have revealed that human-interpretable concepts sometimes emerge as individual latent variables within those networks: for example, object detector units emerge within networks trained to recognize places \cite{zhou2014object}; part detectors emerge in object classifiers \cite{gonzalez2016semantic}; and object detectors emerge in generative video networks \cite{vondrick2016generating} 
(Fig.~\ref{cited-figure}).
This internal structure has appeared in situations where the networks are not constrained to decompose problems in any interpretable way.

The emergence of interpretable structure suggests that deep networks may be learning disentangled representations spontaneously. While it is commonly understood that a network can learn an efficient encoding that makes economical use of hidden variables to distinguish its states, the appearance of a disentangled representation is not well-understood. A disentangled representation aligns its variables with a meaningful factorization of the underlying problem structure, and encouraging disentangled representations is a significant area of research \cite{bengio2013representation}. If the internal representation of a deep network is partly disentangled, one possible path for understanding its mechanisms is to detect disentangled structure, and simply read out the separated factors.
\let\thefootnote\relax\footnotetext{$*$ indicates equal contribution}

However, this proposal raises questions which we address in this paper:

{\begin{itemize}[itemsep=0mm,topsep=0mm,parsep=0mm]
\item What is a disentangled representation, and how can its factors be quantified and detected?
\item Do interpretable hidden units reflect a special alignment of feature space, or are interpretations a chimera?
\item What conditions in state-of-the-art training lead to representations with greater or lesser entanglement?
\end{itemize}
}

To examine these issues, we propose a general analytic framework, \textit{network dissection}, for interpreting deep visual representations and quantifying their interpretability. Using Broden, a broadly and densely labeled data set, our framework identifies hidden units' semantics for any given CNN, then aligns them with human-interpretable concepts. We evaluate our method on various CNNs (AlexNet, VGG, GoogLeNet, ResNet) trained on object and scene recognition, and show that emergent interpretability is an axis-aligned property of a representation that can be destroyed by rotation without affecting discriminative power. We further examine how interpretability is affected by training data sets, training techniques like dropout \cite{srivastava2014dropout} and batch normalization \cite{ioffe2015batch}, and supervision by different primary tasks\footnote{Source code and data available at \url{http://netdissect.csail.mit.edu}}.

\subsection{Related Work}

A growing number of techniques have been developed to understand the internal representations of convolutional neural networks through visualization. The behavior of a CNN can be visualized by sampling image patches that maximize activation of hidden units \cite{zeiler2014visualizing,zhou2014object}, or by using variants of backpropagation to identify or generate salient image features \cite{mahendran2004understanding,simonyan2013deep,zeiler2014visualizing}.  The discriminative power of hidden layers of CNN features can also be understood by isolating portions of networks, transferring them or limiting them, and testing their capabilities on specialized problems \cite{yosinski2014transferable,razavian2014cnn,agrawal2014analyzing}. Visualizations digest the mechanisms of a network down to images which themselves must be interpreted; this motivates our work which aims to match representations of CNNs with labeled interpretations directly and automatically.

Most relevant to our current work are explorations of the roles of individual units inside neural networks.  In~\cite{zhou2014object} human evaluation was used to determine that individual units behave as object detectors in a network that was trained to classify scenes. \cite{nguyen2016synthesizing} automatically generated prototypical images for individual units by learning a feature inversion mapping; this contrasts with our approach of automatically assigning concept labels. Recently \cite{alain2016understanding} suggested an approach to testing the intermediate layers by training simple linear probes, which analyzes the information dynamics among layers and its effect on the final prediction.

\section{Network Dissection}

How can we quantify the clarity of an idea? The notion of a disentangled representation rests on the human perception of what it means for a concept to be mixed up. Therefore when we quantify interpretability, we define it in terms of alignment with a set of human-interpretable concepts. Our measurement of interpretability for deep visual representations proceeds in three steps:
\begin{enumerate}
\item Identify a broad set of human-labeled visual concepts.
\item Gather hidden variables' response to known concepts.
\item Quantify alignment of hidden variable$-$concept pairs.
\end{enumerate}
This three-step process of \emph{network dissection} is reminiscent of the procedures used by neuroscientists to understand similar representation questions in biological neurons~\cite{quiroga2005invariant}. Since our purpose is to measure the level to which a representation is disentangled, we focus on quantifying the correspondence between a single latent variable and a visual concept.

In a fully interpretable local coding such as a one-hot-encoding, each variable will match exactly with one human-interpretable concept. Although we expect a network to learn partially nonlocal representations in interior layers \cite{bengio2013representation}, and past experience shows that an emergent concept will often align with a combination of a several hidden units \cite{gonzalez2016semantic,agrawal2014analyzing}, our present aim is to assess how well a representation is disentangled. Therefore we measure the alignment between single units and single interpretable concepts. This does not gauge the discriminative power of the representation; rather it quantifies its disentangled interpretability. As we will show in Sec.~\ref{rotations}, it is possible for two representations of perfectly equivalent discriminative power to have very different levels of interpretability.

To assess the interpretability of any given CNN, we draw concepts from a new broadly and densely labeled image data set that unifies labeled visual concepts from a heterogeneous collection of labeled data sources, described in Sec.~\ref{section-broden}. We then measure the alignment of each hidden unit of the CNN with each concept by evaluating the feature activation of each individual unit as a segmentation model for each concept. To quantify the interpretability of a layer as a whole, we count the number of distinct visual concepts that are aligned with a unit in the layer, as detailed in Sec.~\ref{section-scoring}.

\subsection{Broden: Broadly and Densely Labeled Dataset}
\label{section-broden}

To be able to ascertain alignment with both low-level concepts such as colors and higher-level concepts such as objects, we have assembled a new heterogeneous data set.
 
\begin{figure}
\begin{center}
{\sffamily\scriptsize\def\arraystretch{1}
\begin{tabularx}{\columnwidth}{@{}X@{}X@{\hspace{.4em}}X@{}X@{\hspace{.4em}}X@{}X@{}}
\multicolumn{2}{@{}l}{street  (scene)}%
& \multicolumn{2}{@{}l}{flower (object)}%
& \multicolumn{2}{@{}l}{headboard (part)} \\
\parbox[c]{1.3cm}{\includegraphics[width=1.3cm,height=1.3cm]{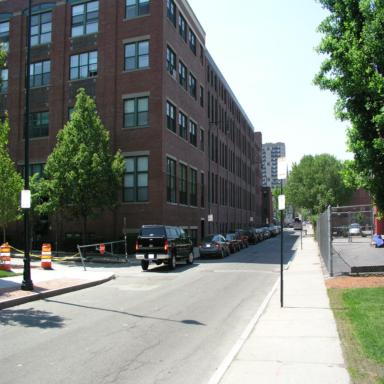}}
& \parbox[c]{1.3cm}{\includegraphics[width=1.3cm,height=1.3cm]{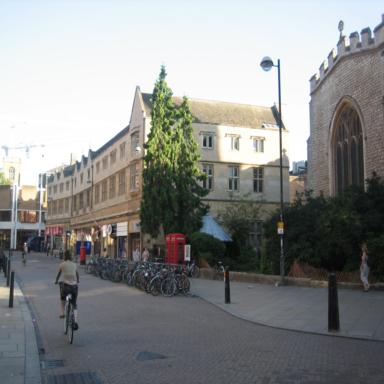}}
& \parbox[c]{1.3cm}{\includegraphics[width=1.3cm,height=1.3cm]{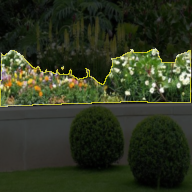}}
& \parbox[c]{1.3cm}{\includegraphics[width=1.3cm,height=1.3cm]{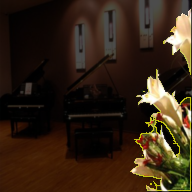}}
& \parbox[c]{1.3cm}{\includegraphics[width=1.3cm,height=1.3cm]{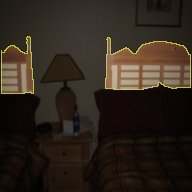}}
& \parbox[c]{1.3cm}{\includegraphics[width=1.3cm,height=1.3cm]{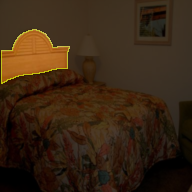}}
\\[9mm]
\multicolumn{2}{@{}l}{swirly (texture)}%
& \multicolumn{2}{@{}l}{pink (color)}%
& \multicolumn{2}{@{}l}{metal (material)} \\
\parbox[c]{1.3cm}{\includegraphics[width=1.3cm,height=1.3cm]{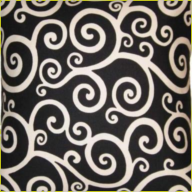}}
& \parbox[c]{1.3cm}{\includegraphics[width=1.3cm,height=1.3cm]{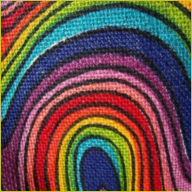}}
& \parbox[c]{1.3cm}{\includegraphics[width=1.3cm,height=1.3cm]{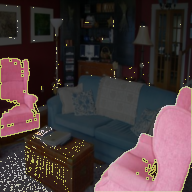}}
& \parbox[c]{1.3cm}{\includegraphics[width=1.3cm,height=1.3cm]{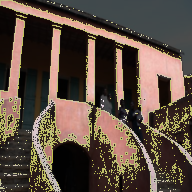}}
& \parbox[c]{1.3cm}{\includegraphics[width=1.3cm,height=1.3cm]{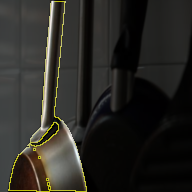}}
& \parbox[c]{1.3cm}{\includegraphics[width=1.3cm,height=1.3cm]{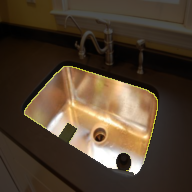}}
\\[9mm]
\end{tabularx}%
}

\end{center}
\vspace{-3mm}
\caption{Samples from the \textbf{Broden} Dataset. The ground truth for each concept is a pixel-wise dense annotation.}\label{sample_broden}
\vspace{-2mm}
\end{figure}

The Broadly and Densely Labeled Dataset (\textbf{Broden}) unifies several densely labeled image data sets: ADE \cite{zhou2016semantic}, OpenSurfaces \cite{bell14intrinsic}, Pascal-Context \cite{mottaghi_cvpr14}, Pascal-Part \cite{chen_cvpr14}, and the Describable Textures Dataset \cite{cimpoi2014describing}. These data sets contain examples of a broad range of objects, scenes, object parts, textures, and materials in a variety of contexts. Most examples are segmented down to the pixel level except textures and scenes which are given for full-images. In addition, every image pixel in the data set is annotated with one of the eleven common color names according to the human perceptions classified by van de Weijer \cite{van2009learning}. A sample of the types of labels in the Broden dataset are shown in Fig.~\ref{sample_broden}.

\begin{table}\caption{Statistics of each label type included in the data set.}
\label{stat_broden}
\centering
\footnotesize
\begin{tabular}{| c | c | c | c |}
  \hline                       
Category & Classes & Sources & Avg sample  \\
    \hline   
scene & 468 & ADE \cite{zhou2016semantic} & 38 \\
object & 584 & ADE \cite{zhou2016semantic}, Pascal-Context \cite{mottaghi_cvpr14} & 491 \\
part   & 234 & ADE \cite{zhou2016semantic}, Pascal-Part \cite{chen_cvpr14} & 854 \\
material & 32 & OpenSurfaces \cite{bell14intrinsic} & 1,703 \\
texture & 47 & DTD \cite{cimpoi2014describing} & 140 \\
color & 11 & Generated & 59,250 \\
\hline
\end{tabular}
\end{table}

The purpose of Broden is to provide a ground truth set of exemplars for a broad set of visual concepts.  The concept labels in Broden are normalized and merged from their original data sets so that every class corresponds to an English word. Labels are merged based on shared synonyms, disregarding positional distinctions such as `left' and `top' and avoiding a blacklist of 29 overly general synonyms (such as `machine' for `car'). Multiple Broden labels can apply to the same pixel: for example, a black pixel that has the Pascal-Part label `left front cat leg' has three labels in Broden: a unified `cat' label representing cats across data sets; a similar unified `leg' label; and the color label `black'.  Only labels with at least 10 image samples are included. Table~\ref{stat_broden} shows the average number of image samples per label class.

\subsection{Scoring Unit Interpretability}
\label{section-scoring}

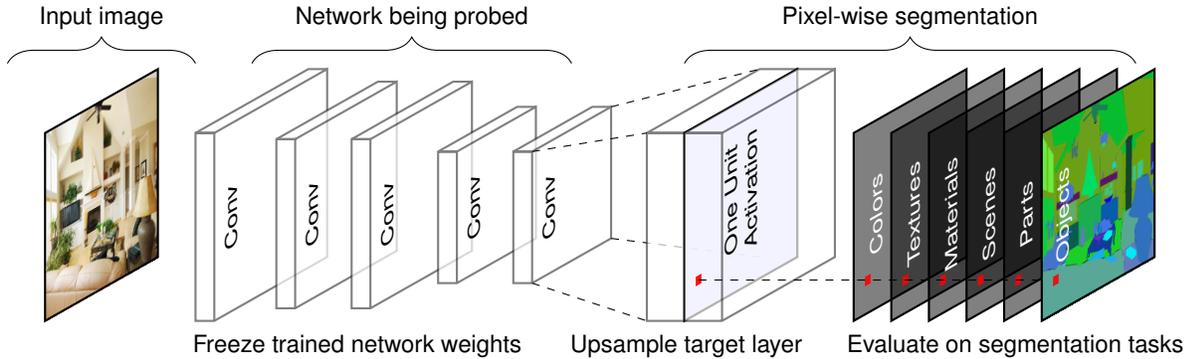
\begin{figure*}
\vspace{-5mm}
\begin{center}
\pgfdeclareimage[height=5cm,width=5cm]{inputimage}{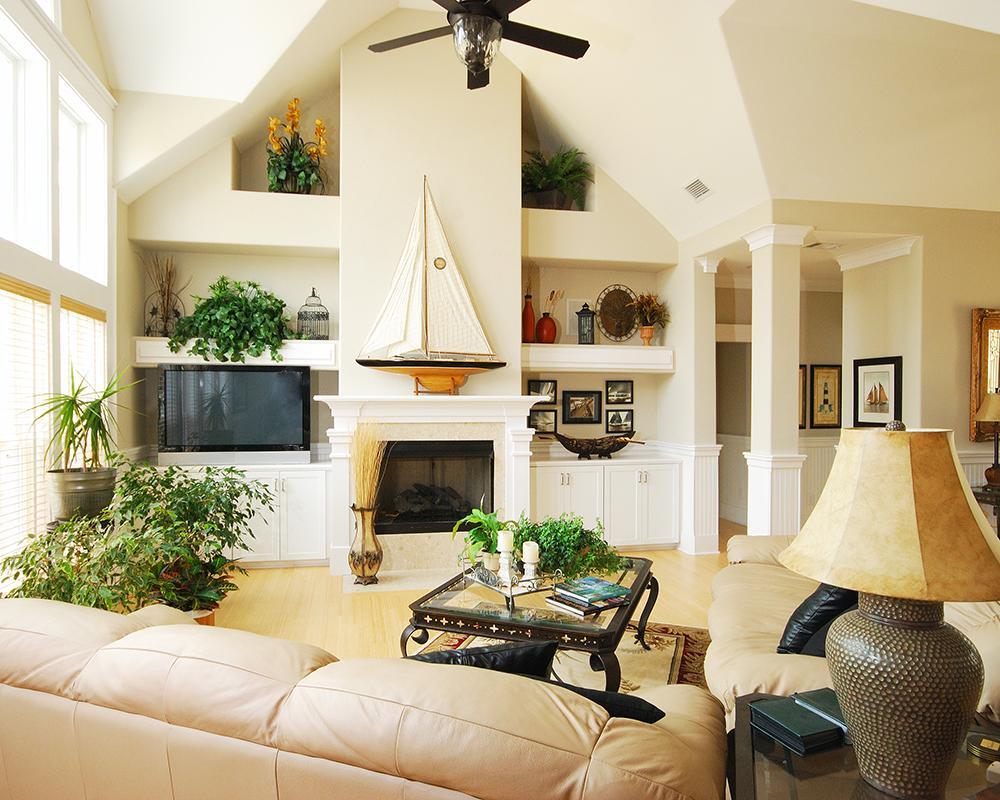}
\pgfdeclareimage[height=5cm,width=5cm]{objectseg}{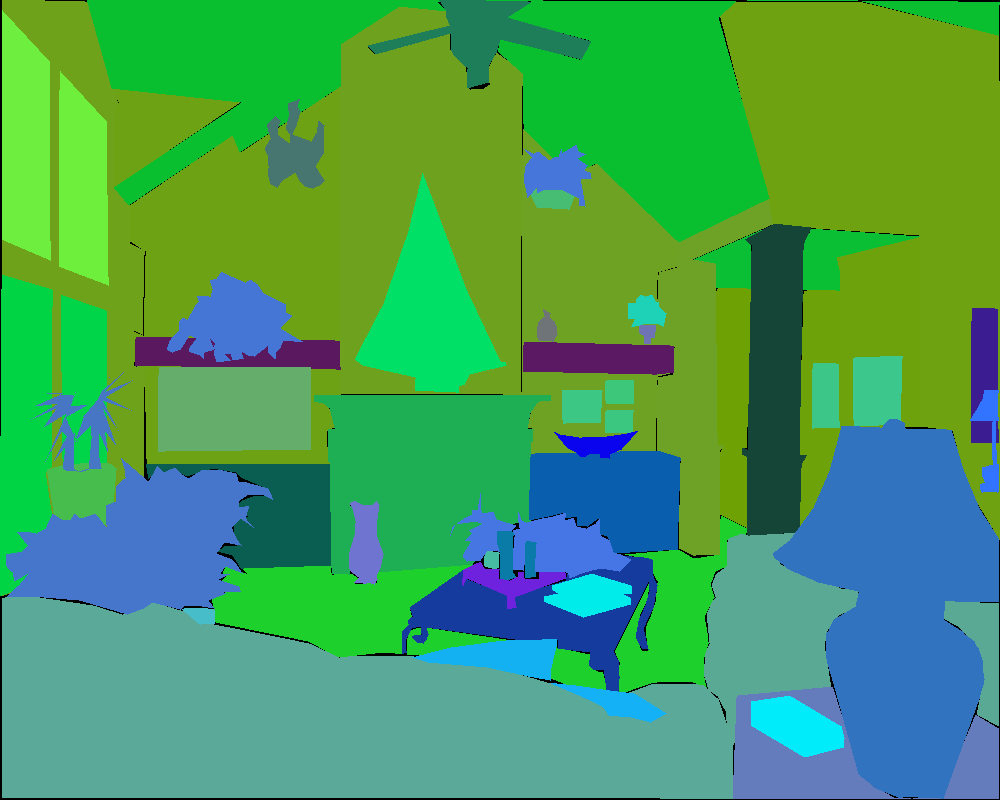}

\makeatletter
\tikzoption{canvas is xy plane at z}[]{%
  \def\tikz@plane@origin{\pgfpointxyz{0}{0}{#1}}%
  \def\tikz@plane@x{\pgfpointxyz{1}{0}{#1}}%
  \def\tikz@plane@y{\pgfpointxyz{0}{1}{#1}}%
  \tikz@canvas@is@plane
}
\def\parsexyz#1#2#3#4{%
    \def\nospace##1{\zap@space##1 \@empty}%
    \def\rawparsexyz(##1,##2,##3){%
        \edef#1{\nospace{##1}}%
        \edef#2{\nospace{##2}}%
        \edef#3{\nospace{##3}}%
    }%
    \expandafter\rawparsexyz#4%
}
\makeatother

\newcommand\slab[2]{
  \parsexyz{\fx}{\fy}{\fz}{#1}
  \parsexyz{\bx}{\by}{\bz}{#2}
    \fill[white,fill opacity=0.7]
        (\bx, \fy, \fz) --
        (\bx, \fy, \bz) --
        (\fx, \fy, \bz) --
        (\fx, \by, \bz) --
        (\fx, \by, \fz) --
        (\bx, \by, \fz) --
        cycle;
  \draw [thick, opacity=0.5]
        (\bx, \fy, \fz) --
        (\bx, \fy, \bz) --
        (\fx, \fy, \bz) --
        (\fx, \by, \bz) --
        (\fx, \by, \fz) --
        (\bx, \by, \fz) --
        (\bx, \fy, \fz) --
        (\fx, \fy, \fz) --
        (\fx, \fy, \bz) --
        (\fx, \fy, \fz) --
        (\fx, \by, \fz);}

\begin{tikzpicture}[y={(0cm,1cm)},x={(0.6cm,0.34cm)}, z={(1cm,0cm)},scale=0.5,every node/.style={font=\sffamily\small}]

\draw [decorate,decoration={brace,amplitude=10pt}] (2.5,6,-7) -- (2.5,6,-2) node [black,midway,yshift=0.6cm] {Input image};
\draw [decorate,decoration={brace,amplitude=10pt}] (2.5,6,-0.5) -- (2.5,6,8) node [black,midway,yshift=0.6cm] {Network being probed};
\draw [decorate,decoration={brace,amplitude=10pt}] (2.5,6,11) -- (2.5,6,23) node [black,midway,yshift=0.6cm] {Pixel-wise segmentation};
\node at (-2,0,5) {Freeze trained network weights};

\begin{scope}[canvas is xy plane at z=-4.5]
    \fill[white,fill opacity=0.9] (0,0) rectangle (5,5);
    \pgftext[at=\pgfpoint{0}{0},left,base]{\pgfuseimage{inputimage}}
    \draw[black,thick] (0,0) rectangle (5,5);
\end{scope}

\slab{(0,5,0)}{(5,0,-0.5)};
\node[canvas is xy plane at z=0,transform shape,
  font=\sffamily\huge,align=center,rotate=90] at (0.75, 2.5)
  {Conv};

\slab{(0.25,4.75,2)}{(4.75,0.25,1.5)};
\node[canvas is xy plane at z=2,transform shape,
  font=\sffamily\huge,align=center,rotate=90] at (1, 2.5)
  {Conv};

\slab{(0.25,4.75,4)}{(4.75,0.25,3.5)};
\node[canvas is xy plane at z=4,transform shape,
  font=\sffamily\huge,align=center,rotate=90] at (1, 2.5)
  {Conv};

\slab{(0.75,4.25,6)}{(4.25,0.75,5.5)};
\node[canvas is xy plane at z=6,transform shape,
  font=\sffamily\huge,align=center,rotate=90] at (1.5, 2.5)
  {Conv};

\slab{(0.75,4.25,8)}{(4.25,0.75,7.5)};
\node[canvas is xy plane at z=8,transform shape,
  font=\sffamily\huge,align=center,rotate=90] at (1.5, 2.5)
  {Conv};

\draw [dashed] (.75,.75,8) -- (0,0,11.5);
\draw [dashed] (4.25,4.25,8) -- (5,5,11.5);
\draw [dashed] (.75,4.25,8) -- (0,5,11.5);
\draw [dashed] (4.25,.75,8) -- (5,0,11.5);
\node at (-2,0,13.75) {Upsample target layer};

\slab{(0,5,12.5)}{(5,0,11.5)};
\begin{scope}[canvas is xy plane at z=12.5]
        \fill[blue!30,fill opacity=0.5] (0,0) rectangle (5,5);
        \draw[black,thick] (0,0) rectangle (5,5);
        \fill[red, fill opacity=0.9] (0.5,0.75) rectangle (.75,1);
\end{scope}

\slab{(0,5,13.5)}{(5,0,12.5)};

\begin{scope}[canvas is xy plane at z=12.5]
    \fill[red, fill opacity=0.9] (0.5,0.75) rectangle (.75,1);
    \node[transform shape,
    font=\sffamily\huge,align=center,rotate=90] at (2.5, 2.5)
    {One Unit\\ Activation};
\end{scope}

\begin{scope}[canvas is xy plane at z=17]
        \fill[black,fill opacity=0.5] (0,0) rectangle (5,5);
        \draw[black,thick] (0,0) rectangle (5,5);
        \node[style={transform shape},font=\sffamily\huge,rotate=90,align=center,color=white] at (1, 2.5) {Colors};
        \fill[red, fill opacity=0.9] (0.5,0.75) rectangle (.75,1);
\end{scope}

\begin{scope}[canvas is xy plane at z=18]
        \fill[black,fill opacity=0.5] (0,0) rectangle (5,5);
        \draw[black,thick] (0,0) rectangle (5,5);
        \node[style={transform shape},font=\sffamily\huge,rotate=90,align=center,color=white] at (1, 2.5) {Textures};
        \fill[red, fill opacity=0.9] (0.5,0.75) rectangle (.75,1);
\end{scope}

\begin{scope}[canvas is xy plane at z=19]
        \fill[black,fill opacity=0.5] (0,0) rectangle (5,5);
        \draw[black,thick] (0,0) rectangle (5,5);
        \node[style={transform shape},font=\sffamily\huge,rotate=90,align=center,color=white] at (1, 2.5) {Materials};
        \fill[red, fill opacity=0.9] (0.5,0.75) rectangle (.75,1);
\end{scope}

\begin{scope}[canvas is xy plane at z=20]
        \fill[black,fill opacity=0.5] (0,0) rectangle (5,5);
        \draw[black,thick] (0,0) rectangle (5,5);
        \node[style={transform shape},font=\sffamily\huge,rotate=90,align=center,color=white] at (1, 2.5) {Scenes};
        \fill[red, fill opacity=0.9] (0.5,0.75) rectangle (.75,1);
\end{scope}

\begin{scope}[canvas is xy plane at z=21]
        \fill[black,fill opacity=0.5] (0,0) rectangle (5,5);
        \draw[black,thick] (0,0) rectangle (5,5);
        \node[style={transform shape},font=\sffamily\huge,rotate=90,align=center,color=white] at (1, 2.5) {Parts};
        \fill[red, fill opacity=0.9] (0.5,0.75) rectangle (.75,1);
\end{scope}

\draw [dashed] (.625,.875,12.5) -- (.625,.875,22);
\node at (-2,0,22.5) {Evaluate on segmentation tasks};

\begin{scope}[canvas is xy plane at z=22]
    \fill[white,fill opacity=0.9] (0,0) rectangle (5,5);
    \pgftext[at=\pgfpoint{0}{0},left,base]{\pgfuseimage{objectseg}}
    \draw[black,thick] (0,0) rectangle (5,5);
    \node[style={transform shape},font=\sffamily\huge,
        rotate=90,align=center,color=white] at (1, 2.5)
        {Objects};
    \fill[red, fill opacity=0.9] (0.5,0.75) rectangle (.75,1);
\end{scope}

\end{tikzpicture}
\end{center}
\vspace{-3mm}
\caption{Illustration of network dissection for measuring semantic alignment of units in a given CNN. Here one unit of the last convolutional layer of a given CNN is probed by evaluating its performance on 1197 segmentation tasks. Our method can probe any convolutional layer.}
\label{learning_framework}
\vspace{-3mm}
\end{figure*}

The proposed network dissection method evaluates every individual convolutional unit in a CNN as a solution to a binary segmentation task to every visual concept in Broden (Fig.~\ref{learning_framework}). Our method can be applied to any CNN using a forward pass without the need for training or backpropagation.

For every input image $\textbf{x}$ in the Broden dataset, the activation map $A_k(\textbf{x})$ of every internal convolutional unit $k$ is collected. Then the distribution of individual unit activations $a_{k}$ is computed. For each unit $k$, the top quantile level $T_k$ is determined such that $P(a_{k} > T_k) = 0.005$ over every spatial location of the activation map in the data set.

To compare a low-resolution unit's activation map to the input-resolution annotation mask $L_{c}$ for some concept $c$, the activation map is scaled up to the mask resolution $S_k(\textbf{x})$ from $A_k(\textbf{x})$ using bilinear interpolation, anchoring interpolants at the center of each unit's receptive field.

$S_k(\textbf{x})$ is then thresholded into a binary segmentation: $M_k(\textbf{x}) \equiv S_k(\textbf{x}) \geq T_k$, selecting all regions for which the activation exceeds the threshold $T_k$.  These segmentations are evaluated against every concept $c$ in the data set by computing intersections $M_k(\textbf{x})\cap L_c(\textbf{x})$, for every $(k, c)$ pair.

The score of each unit $k$ as segmentation for concept $c$ is reported as a data-set-wide intersection over union score
\begin{equation}
IoU_{k,c} = \frac{\sum | M_k(\textbf{x})\cap L_c(\textbf{x})|}{\sum | M_k(\textbf{x})\cup L_c(\textbf{x})|},
\end{equation}
where $|\cdot|$ is the cardinality of a set. Because the data set contains some types of labels which are not present on some subsets of inputs, the sums are computed only on the subset of images that have at least one labeled concept of the same category as $c$. The value of $IoU_{k,c}$ is the accuracy of unit $k$ in detecting concept $c$; we consider one unit $k$ as a detector for concept $c$ if $IoU_{k,c}$ exceeds a threshold. Our qualitative results are insensitive to the IoU threshold: different thresholds denote different numbers of units as concept detectors across all the networks but relative orderings remain stable. For our comparisons we report a detector if $IoU_{k,c} > 0.04$. Note that one unit might be the detector for multiple concepts; for the purpose of our analysis, we choose the top ranked label. To quantify the interpretability of a layer, we count the number unique concepts aligned with units. We call this the number of \textit{unique detectors}.

The IoU evaluating the quality of the segmentation of a unit is an objective confidence score for interpretability that is \textit{comparable across networks}. Thus this score enables us to compare interpretability of different representations and lays the basis for the experiments below. Note that network dissection works only as well as the underlying data set: if a unit matches a human-understandable concept that is absent in Broden, then it will not score well for interpretability. Future versions of Broden will be expanded to include more kinds of visual concepts.

\section{Experiments}
\label{sec-experiments}
\begin{table}\caption{Tested CNNs Models}
\label{modelzoo}
\centering
\footnotesize
\begin{tabular}{ccl}
\hline
\textbf{Training} & \textbf{Network} & \textbf{Data set or task}\\
\hline
none & AlexNet & random \\ 
\hline
\multirow{4}*{Supervised} & AlexNet & ImageNet, Places205, Places365, Hybrid. \\
 & GoogLeNet & ImageNet, Places205, Places365. \\
 & VGG-16 & ImageNet, Places205, Places365, Hybrid.  \\
 & ResNet-152 & ImageNet, Places365. \\
 \hline
\multirow{4}*{Self} & \multirow{4}*{AlexNet} & \texttt{context}, \texttt{puzzle}, \texttt{egomotion}, \\
&  & \texttt{tracking}, \texttt{moving}, \texttt{videoorder},\\
& &\texttt{audio}, \texttt{crosschannel},\texttt{colorization}. \\
& & \texttt{objectcentric}. \\
\hline
\end{tabular}
\vspace{-4mm}
\end{table}

For testing we prepare a collection of CNN models with different network architectures and supervision of primary tasks, as listed in Table~\ref{modelzoo}. The network architectures include AlexNet \cite{krizhevsky2012imagenet}, GoogLeNet \cite{szegedy2015going}, VGG \cite{simonyan2014very}, and ResNet \cite{he2016deep}. For supervised training, the models are trained from scratch (i.e., not pretrained) on ImageNet \cite{russakovsky2015imagenet}, Places205 \cite{zhou2014learning}, and Places365 \cite{zhou2016places}. ImageNet is an object-centric data set, which contains 1.2 million images from 1000 classes. Places205 and Places365 are two subsets of the Places Database, which is a scene-centric data set with categories such as kitchen, living room, and coast. Places205 contains 2.4 million images from 205 scene categories, while Places365 contains 1.6 million images from 365 scene categories. ``Hybrid'' refers to a combination of ImageNet and Places365. For self-supervised training tasks, we select several recent models trained on predicting context (\texttt{context}) \cite{doersch2015unsupervised}, solving puzzles (\texttt{puzzle}) \cite{noroozi2016unsupervised}, predicting ego-motion (\texttt{egomotion})~\cite{jayaraman2015learning}, learning by moving (\texttt{moving})~\cite{agrawal2015learning}, predicting video frame order (\texttt{videoorder})~\cite{misra2016shuffle} or tracking (\texttt{tracking})~\cite{wang2015unsupervised}, detecting object-centric alignment (\texttt{objectcentric}) \cite{gao2016object}, colorizing images (\texttt{colorization})~\cite{zhang2016colorful}, predicting cross-channel (\texttt{crosschannel})~\cite{zhang2016splitbrain}, and predicting ambient sound from frames (\texttt{sound})~\cite{owens2016ambient}. The self-supervised models we analyze are comparable to each other in that they all use AlexNet or an AlexNet-derived architecture. 

In the following experiments, we begin by validating our method using human evaluation. Then, we use random unitary rotations of a learned representation to test whether interpretability of CNNs is an axis-independent property; we find that it is not, and we conclude that interpretability is not an inevitable result of the discriminative power of a representation.  Next, we analyze all the convolutional layers of AlexNet as trained on ImageNet \cite{krizhevsky2012imagenet} and as trained on Places \cite{zhou2014learning}, and confirm that our method reveals detectors for higher-level concepts at higher layers and lower-level concepts at lower layers; and that more detectors for higher-level concepts emerge under scene training. Then, we show that different network architectures such as AlexNet, VGG, and ResNet yield different interpretability, while differently supervised training tasks and self-supervised training tasks also yield a variety of levels of interpretability. Finally we show the impact of different training conditions, examine the relationship between discriminative power and interpretability, and investigate a possible way to improve the interpretability of CNNs by increasing their width.

\subsection{Human Evaluation of Interpretations}
\label{section-human}

We evaluate the quality of the unit interpretations found by our method using Amazon Mechanical Turk (AMT). Raters were shown 15 images with highlighted patches showing the most highly-activating regions for each unit in AlexNet trained on Places205, and asked to decide (yes/no) whether a given phrase describes most of the image patches.

Table~\ref{comparison_quantitative} summarizes the results. First, we determined the set of interpretable units as those units for which raters agreed with ground-truth interpretations from \cite{zhou2014object}. Over this set of units, we report the portion of interpretations generated by our method that were rated as descriptive. Within this set we also compare to the portion of ground-truth labels that were found to be descriptive by a second group of raters. The proposed method can find semantic labels for units that are comparable to descriptions written by human annotators at the highest layer. At the lowest layer, the low-level color and texture concepts available in Broden are only sufficient to match good interpretations for a minority of units. Human consistency is also highest at \texttt{conv5}, which suggests that humans are better at recognizing and agreeing upon high-level visual concepts such as objects and parts, rather than the shapes and textures that emerge at lower layers.

\begin{table}\caption{Human evaluation of our Network Dissection approach. Interpretable units are those where raters agreed with ground-truth interpretations. Within this set we report the portion of interpretations assigned by our method that were rated as descriptive. Human consistency is based on a second evaluation of ground-truth labels.}
\label{comparison_quantitative}
\centering
\footnotesize
\begin{tabular}{@{\hspace{1mm}}lll@{\hspace{2.2mm}}l@{\hspace{2.2mm}}l@{\hspace{2.2mm}}l@{\hspace{1mm}}}
\hline
 & conv1 & conv2 & conv3 & conv4 & conv5 \\
\hline
Interpretable units & 57/96 & 126/256 & 247/384 & 258/384 & 194/256 \\
\hline
Human consistency & 82\% & 76\% & 83\% & 82\% & 91\% \\
Network Dissection & 37\% & 56\% & 54\% & 59\% & 71\% \\
\hline
\end{tabular}
\vspace{-3mm}
\end{table}

\subsection{Measurement of Axis-Aligned Interpretability}
\label{rotations}

We conduct an experiment to determine whether it is meaningful to assign an interpretable concept to an individual unit. Two possible hypotheses can explain the emergence of interpretability in individual hidden layer units:
\begin{enumerate}[itemindent=4em]
\item[Hypothesis 1.] Interpretable units emerge because interpretable concepts appear in most directions in representation space. If the representation localizes related concepts in an axis-independent way, projecting to \textit{any} direction could reveal an interpretable concept, and interpretations of single units in the natural basis may not be a meaningful way to understand a representation.
\item[Hypothesis 2.] Interpretable alignments are unusual, and interpretable units emerge because learning converges to a special basis that aligns explanatory factors with individual units. In this model, the natural basis represents a meaningful decomposition learned by the network.
\end{enumerate}
Hypothesis 1 is the default assumption: in the past it has been found~\cite{szegedy2013intriguing} that with respect to interpretability ``there is no distinction between individual high level units and random linear combinations of high level units.''

Network dissection allows us to re-evaluate this hypothesis. We apply random changes in basis to a representation learned by AlexNet. Under hypothesis 1, the overall level of interpretability should not be affected by a change in basis, even as rotations cause the specific set of represented concepts to change. Under hypothesis 2, the overall level of interpretability is expected to drop under a change in basis.

\begin{figure}
\begin{center}
\vspace{-3mm}
\includegraphics[width=1\linewidth,trim={3mm 10mm 3mm -3mm}]{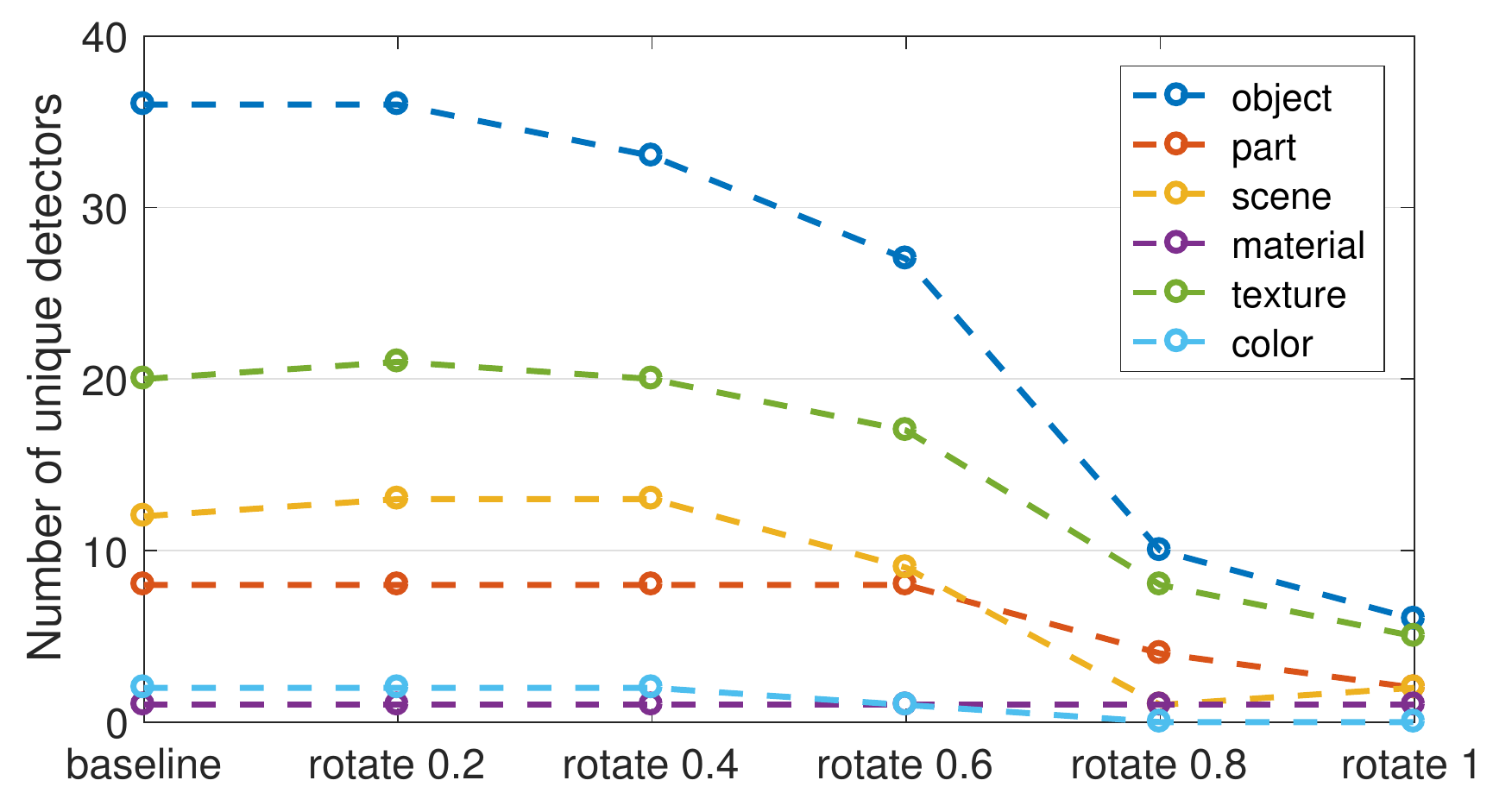}
\vspace{-3mm}
\end{center}
\caption{Interpretability over changes in basis of the representation of AlexNet \texttt{conv5} trained on Places. The vertical axis shows the number of unique interpretable concepts that match a unit in the representation.  The horizontal axis shows $\alpha$, which quantifies the degree of rotation.}
\label{rotation-plot}
\end{figure}

We begin with the representation of the 256 convolutional units of AlexNet \texttt{conv5} trained on Places205 and examine the effect of a change in basis. To avoid any issues of conditioning or degeneracy, we change basis using a random orthogonal transformation $Q$. The rotation $Q$ is drawn uniformly from $SO(256)$ by applying Gram-Schmidt on a normally-distributed $QR = A \in \mathbf{R}^{256^2}$ with positive-diagonal right-triangular $R$, as described by \cite{diaconis2005random}. Interpretability is summarized as the number of unique visual concepts aligned with units, as defined in Sec.~\ref{section-scoring}.

Denoting AlexNet \texttt{conv5} as $f(x)$, we find that the number of unique detectors in $Qf(x)$ is 80\% fewer than the number of unique detectors in $f(x)$. Our finding is inconsistent with hypothesis 1 and consistent with hypothesis 2.

We also test smaller perturbations of basis using $Q^{\alpha}$ for $0 \leq \alpha \leq 1$,
where the fractional powers $Q^\alpha \in SO(256)$ are chosen to form a minimal geodesic gradually rotating from $I$ to $Q$; these intermediate rotations are computed using a Schur decomposition. Fig.~\ref{rotation-plot} shows that interpretability of $Q^\alpha f(x)$ decreases as larger rotations are applied.

Each rotated representation has exactly the same discriminative power as the original layer. Writing the original network as $g(f(x))$, note that $g'(r) \equiv g(Q^T r)$ defines a neural network that processes the rotated representation $r = Q f(x)$ exactly as the original $g$ operates on $f(x)$.  We conclude that interpretability is neither an inevitable result of discriminative power, nor is it a prerequisite to discriminative power.  Instead, we find that interpretability is a different quality that must be measured separately to be understood.

\begin{figure*}
\vspace{-2mm}
\begin{center}
\newcommand{\sideheader}[1]{
\begin{tikzpicture}
    [font=\sffamily,every node/.style={inner sep=0,outer sep=0}]
    \draw[yshift=0.6cm,font=\sffamily] node
        [right=0cm,minimum width=4.2cm,minimum height=0.6cm,
          fill=black!10,rotate=90]
        {#1};
\end{tikzpicture}%
}

\newcommand{\smallimageheader}[1]{
\begin{tikzpicture}
    [font=\sffamily,every node/.style={inner sep=0,outer sep=0}]
    \draw[yshift=0.6cm,font=\sffamily\scriptsize]
        node
        [right=0cm,minimum width=2.39cm,minimum height=0.4cm,
          fill=black!10]
        {#1};
\end{tikzpicture}%
}
\newcommand{\smallimageblock}[1]{
\begin{tikzpicture}
    [font=\sffamily,every node/.style={inner sep=0,outer sep=0}]
    \foreach \ti / \trbl / \tlbl / \tdir / \timg in {#1}
    {
        \pgfmathsetmacro\tya{\ti * 1.13-.52}
        \pgfmathsetmacro\tyb{\ti * 1.13}
        \draw[yshift=-{\tya}cm]
        node[right=0cm,
             text width=2.39cm,
             minimum width=2.39cm,minimum height=0.5cm,
             align=left,font=\sffamily\tiny] {\tlbl}
        node[right=0cm,
             text width=2.39cm,
             minimum width=2.39cm,minimum height=0.5cm,
             align=right,font=\sffamily\tiny] {\trbl};
        \draw[yshift=-{\tyb}cm]
		node[right=0cm,align=left]{\includegraphics
            [trim=0 0 225px 0,clip,height=0.8cm,width=2.39cm]
            {images/\tdir/\timg.jpg}%
        };
    }
\end{tikzpicture}%
}

\begin{tabular}{@{}
  c@{\hspace{1mm}}
  c@{\hspace{1mm}}
  c@{\hspace{1mm}}
  c@{\hspace{1mm}}
  c@{\hspace{1mm}}
  c@{}}
\includegraphics[width=5cm,trim={3mm 1.1cm 0 0}]{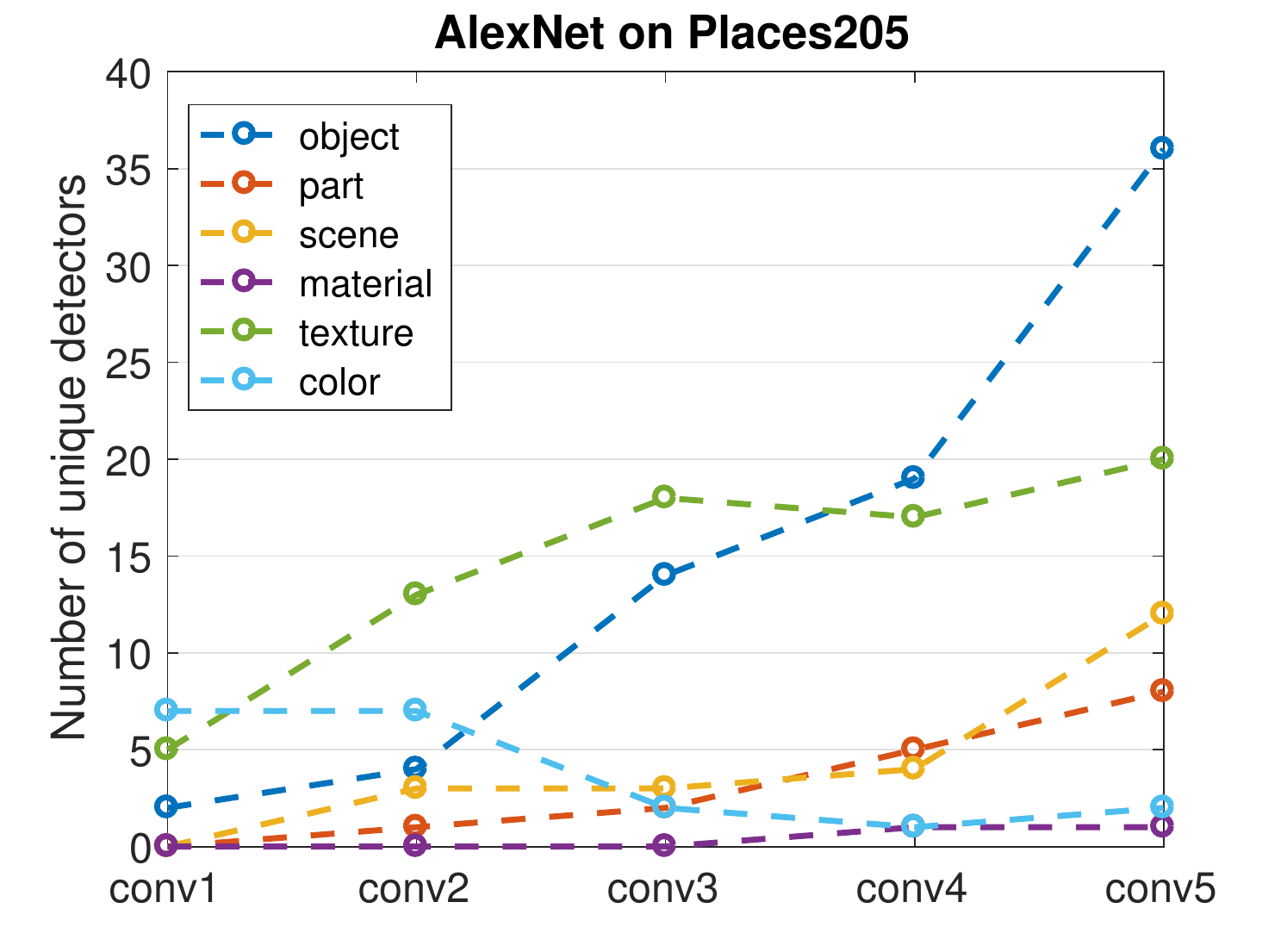} &%
\smallimageblock{
0/h:green/veined (texture)/places205/conv1-0043-veined,
1/h:color yellow/orange (color)/places205/conv1-0055-orange,
2/h:pink or red/red (color)/places205/conv1-0071-red%
} &
\smallimageblock{
0/h:sky/sky (object)/places205/conv2-0207-sky,
1/h:black\&white/lacelike (texture)/places205/conv2-0248-lacelike,
2/h:grid pattern/lined (texture)/places205/conv2-0127-lined%
} &
\smallimageblock{
0/h:grass/grass (object)/places205/conv3-0127-grass,
1/h:corrugated/banded (texture)/places205/conv3-0254-banded,
2/h:pattern/perforated (texture)/places205/conv3-0300-perforated%
} &
\smallimageblock{
0/h:windows/chequered (texture)/places205/conv4-0022-chequered,
1/h:tree/tree (object)/places205/conv4-0014-tree,
2/h:horiz. lines/crosswalk (part)/places205/conv4-0087-crosswalk%
} &
\smallimageblock{
0/h:bed/bed (object)/places205/conv5-0115-bed,
1/h:car/car (object)/places205/conv5-0086-car,
2/h:montain/mountain (scene)/places205/conv5-0139-mountain-snowy%
} \\[2mm]
\includegraphics[width=5cm,trim={3mm 1.1cm 0 0}]{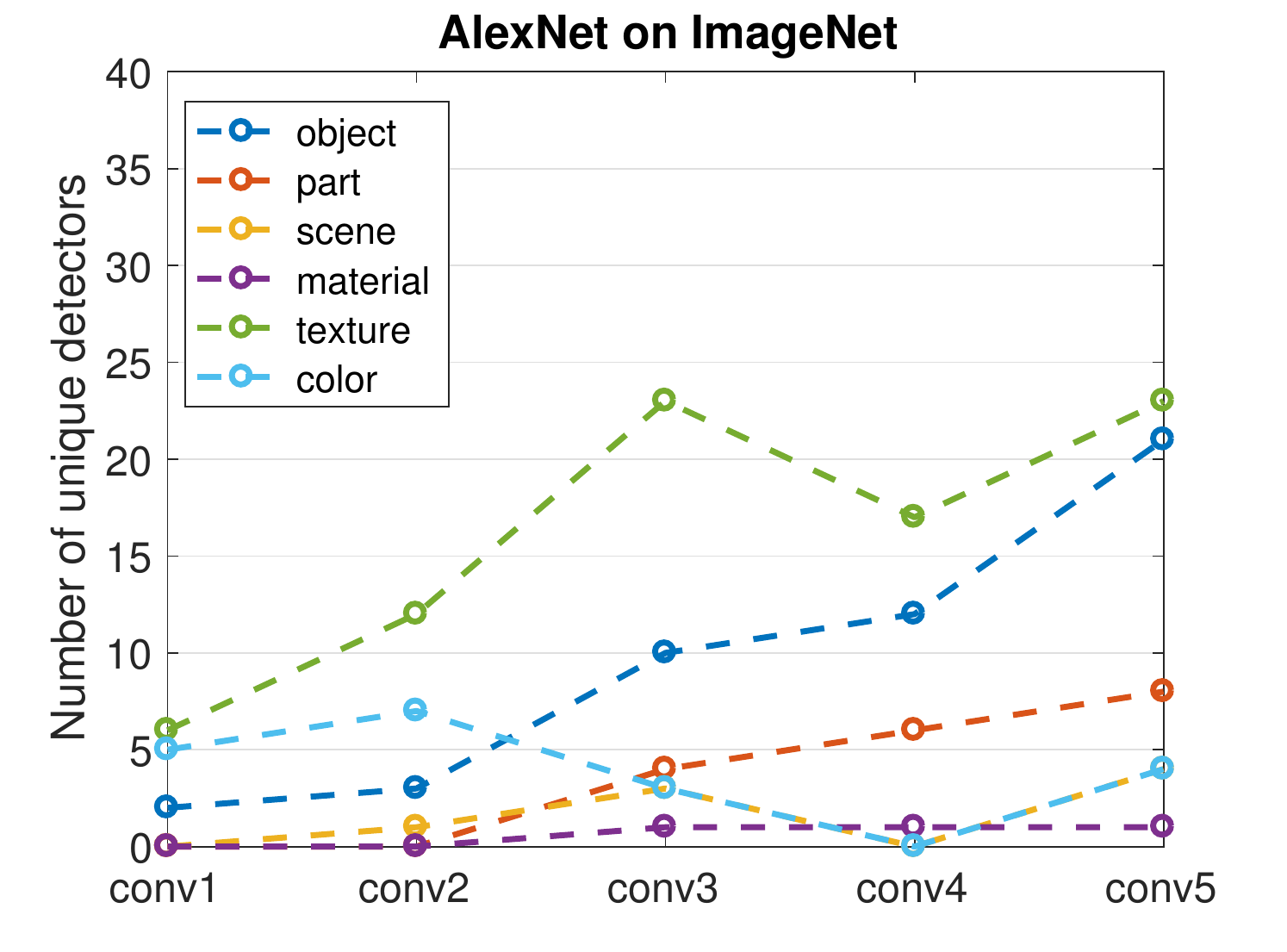} &%
\smallimageblock{
0/h:red/red (color)/imagenet2/conv1-0065-red-red,
1/h:yellow/yellow (color)/imagenet2/conv1-0085-yellow-yellow,
2/h:blue/sky (object)/imagenet2/conv1-0057-blue-sky%
} &
\smallimageblock{
0/h:yellow/woven (texture)/imagenet2/conv2-0073-striped-pattern-lines,
1/h:striped/banded (texture)/imagenet2/conv2-0162-monochrome-lacelike,
2/h:mesh/grid (texture)/imagenet2/conv2-0193-colorful-shapes-zigzagged%
} &
\smallimageblock{
0/h:orange/food (material)/imagenet2/conv3-0031-orange-food,
1/h:blue sky/sky (object)/imagenet2/conv3-0051-blue-sky-sky,
2/h:nosed/dotted (texture)/imagenet2/conv3-0240-noses-dotted%
} &
\smallimageblock{
0/h:animal face/muzzle (part)/imagenet2/conv4-0371-animal-face-muzzle,
1/h:round/swirly (texture)/imagenet2/conv4-0246-round-swirly,
2/h:face/head (part)/imagenet2/conv4-0368-face-head%
} &
\smallimageblock{
0/h:wheels/wheel (part)/imagenet2/conv5-0197-wheels-wheel,
1/h:animal faces/cat (object)/imagenet2/conv5-0154-animal-faces-cat,
2/h:leg/leg (part)/imagenet2/conv5-0192-leg-leg%
} \\
& \smallimageheader{conv1}
& \smallimageheader{conv2}
& \smallimageheader{conv3}
& \smallimageheader{conv4}
& \smallimageheader{conv5}
\\
\end{tabular}

\end{center}
\vspace{-3mm}
\caption{A comparison of the interpretability of all five convolutional layers of AlexNet, as trained on classification tasks for Places (top) and ImageNet (bottom). At right, three examples of units in each layer are shown with identified semantics. The segmentation generated by each unit is shown on the three Broden images with highest activation. Top-scoring labels are shown above to the left, and human-annotated labels are shown above to the right. Some disagreement can be seen for the dominant judgment of meaning. For example, human annotators mark the first \texttt{conv4} unit on Places as a `windows' detector, while the algorithm matches the `chequered' texture.}\label{places_imagenet}
\end{figure*}
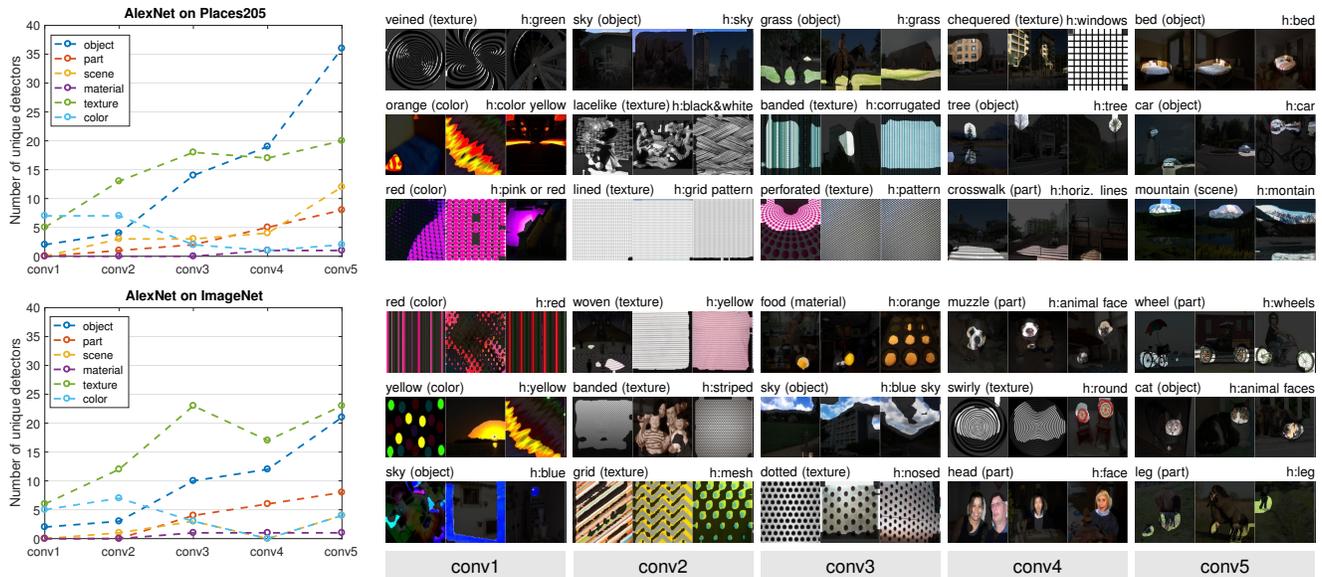

\begin{figure*}
\begin{center}
\includegraphics[width=\textwidth]{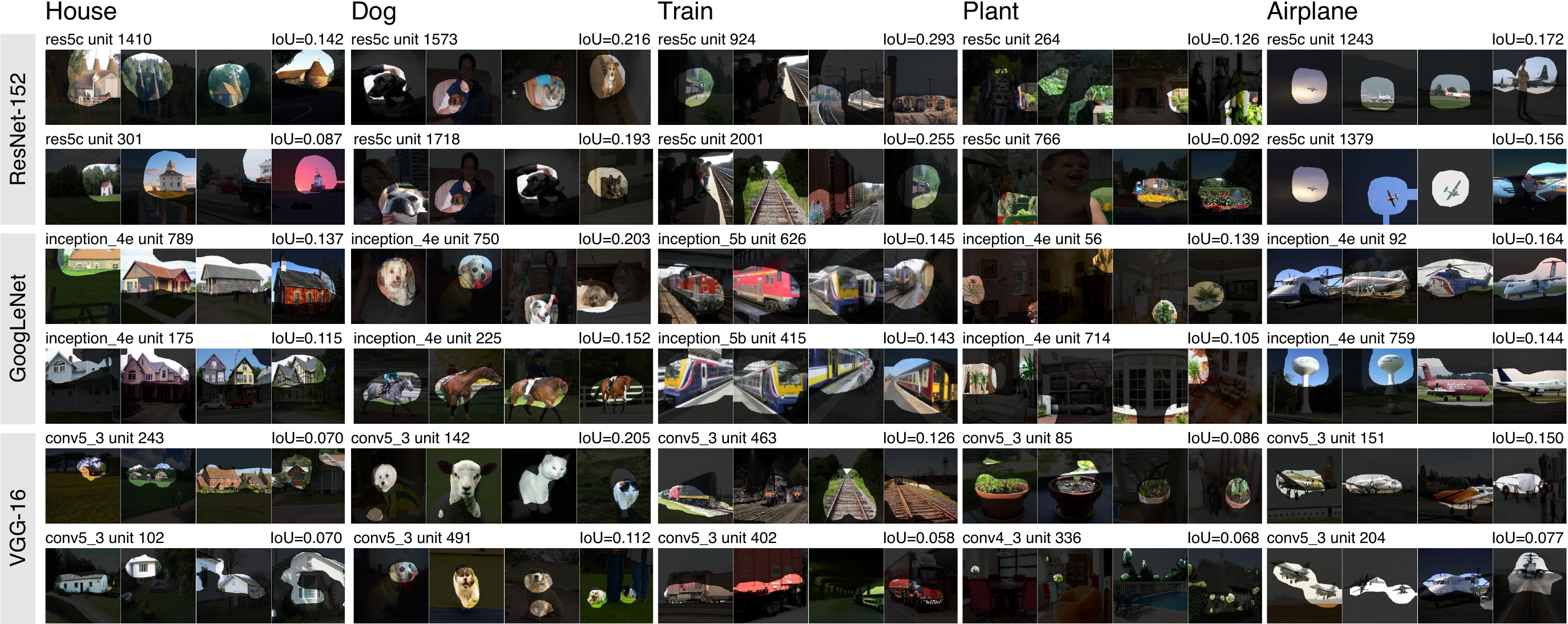}
\end{center}
\vspace{-3mm}
\caption{A comparison of several visual concept detectors identified by network dissection in ResNet, GoogLeNet, and VGG. Each network is trained on Places365. The two highest-IoU matches among convolutional units of each network is shown. The segmentation generated by each unit is shown on the four maximally activating Broden images. Some units activate on concept generalizations, e.g., GoogLeNet 4e's unit 225 on horses and dogs, and 759 on white ellipsoids and jets.}
\label{figure-concepts}
\vspace{-3mm}
\end{figure*}

\subsection{Disentangled Concepts by Layer}

Using network dissection, we analyze and compare the interpretability of units within all the convolutional layers of Places-AlexNet and ImageNet-AlexNet. Places-AlexNet is trained for scene classification on Places205 \cite{zhou2014learning}, while ImageNet-AlexNet is the identical architecture trained for object classification on ImageNet \cite{krizhevsky2012imagenet}.

The results are summarized in Fig.~\ref{places_imagenet}. A sample of units are shown together with both automatically inferred interpretations and manually assigned interpretations taken from \cite{zhou2014object}. We can see that the predicted labels match the human annotation well, though sometimes they capture a different description of a visual concept, such as the `crosswalk' predicted by the algorithm compared to `horizontal lines' given by the human for the third unit in \texttt{conv4} of Places-AlexNet in Fig.~\ref{places_imagenet}. Confirming intuition, color and texture concepts dominate at lower layers \texttt{conv1} and \texttt{conv2} while more object and part detectors emerge in \texttt{conv5}.

\subsection{Network Architectures and Supervisions}
\label{section-comparing-architecture}

\begin{figure}
\vspace{-4mm}
\begin{center}
\includegraphics[width=1\linewidth]{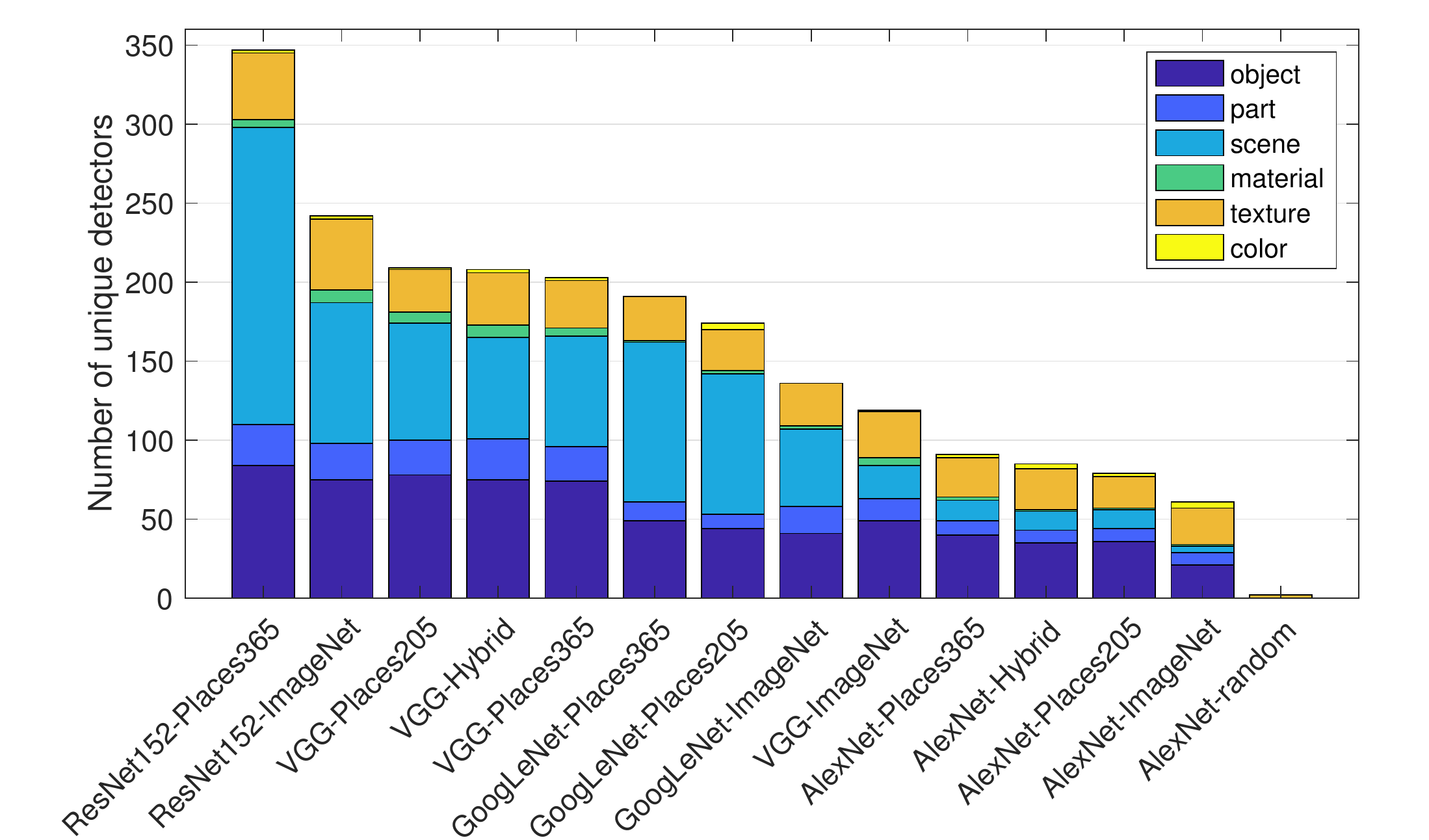}
\end{center}
\vspace{-4mm}
\caption{Interpretability across different architectures and training. }\label{architecture}
\end{figure}

\begin{figure}
\vspace{-1mm}
\begin{center}
\includegraphics[width=1\linewidth]{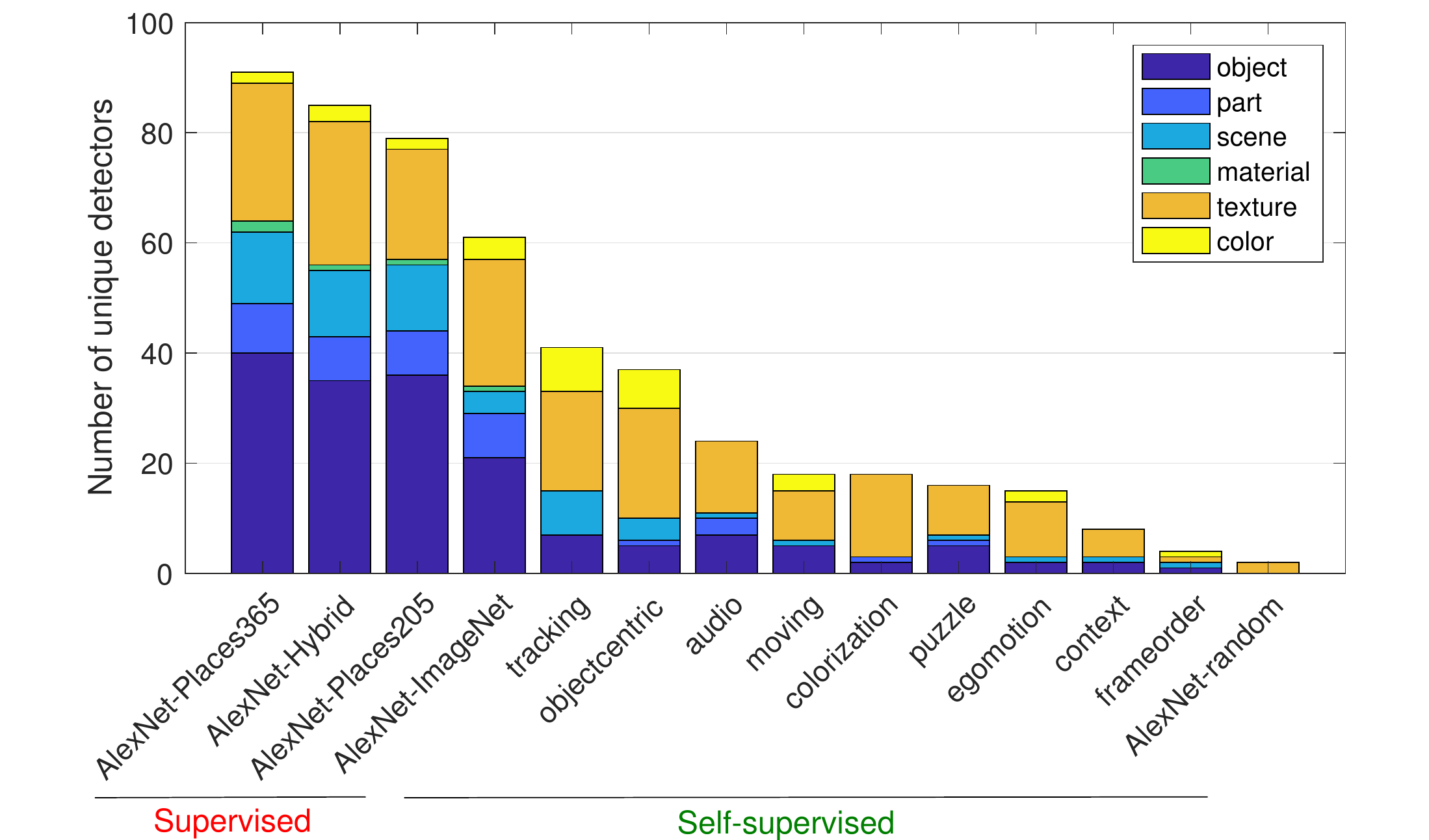}
\end{center}
\vspace{-4mm}
\caption{Semantic detectors emerge across different supervision of the primary training task. All these models use the AlexNet architecture and are tested at \texttt{conv5}.}\label{supervision}
\end{figure}

How do different network architectures and training supervisions affect disentangled interpretability of the learned representations? We apply network dissection to evaluate a range of network architectures and supervisions. For simplicity, the following experiments focus on the last convolutional layer of each CNN, where semantic detectors emerge most.

Results showing the number of unique detectors that emerge from various network architectures trained on ImageNet and Places are plotted in Fig.~\ref{architecture}, with examples shown in Fig.~\ref{figure-concepts}. In terms of network architecture, we find that interpretability of ResNet $>$ VGG $>$ GoogLeNet $>$ AlexNet. Deeper architectures appear to allow greater interpretability. Comparing training data sets, we find Places $>$ ImageNet. As discussed in \cite{zhou2014object}, one scene is composed of multiple objects, so it may be beneficial for more object detectors to emerge in CNNs trained to recognize scenes.

Results from networks trained on various supervised and self-supervised tasks are shown in Fig.~\ref{supervision}. Here the network architecture is AlexNet for each model,  We observe that training on Places365 creates the largest number of unique detectors. Self-supervised models create many texture detectors but relatively few object detectors; apparently, supervision from a self-taught primary task is much weaker at inferring interpretable concepts than supervised training on a large annotated data set. The form of self-supervision makes a difference: for example, the colorization model is trained on colorless images, and almost no color detection units emerge. We hypothesize that emergent units represent concepts required to solve the primary task. 

\begin{figure}
\includegraphics[width=0.48\textwidth]{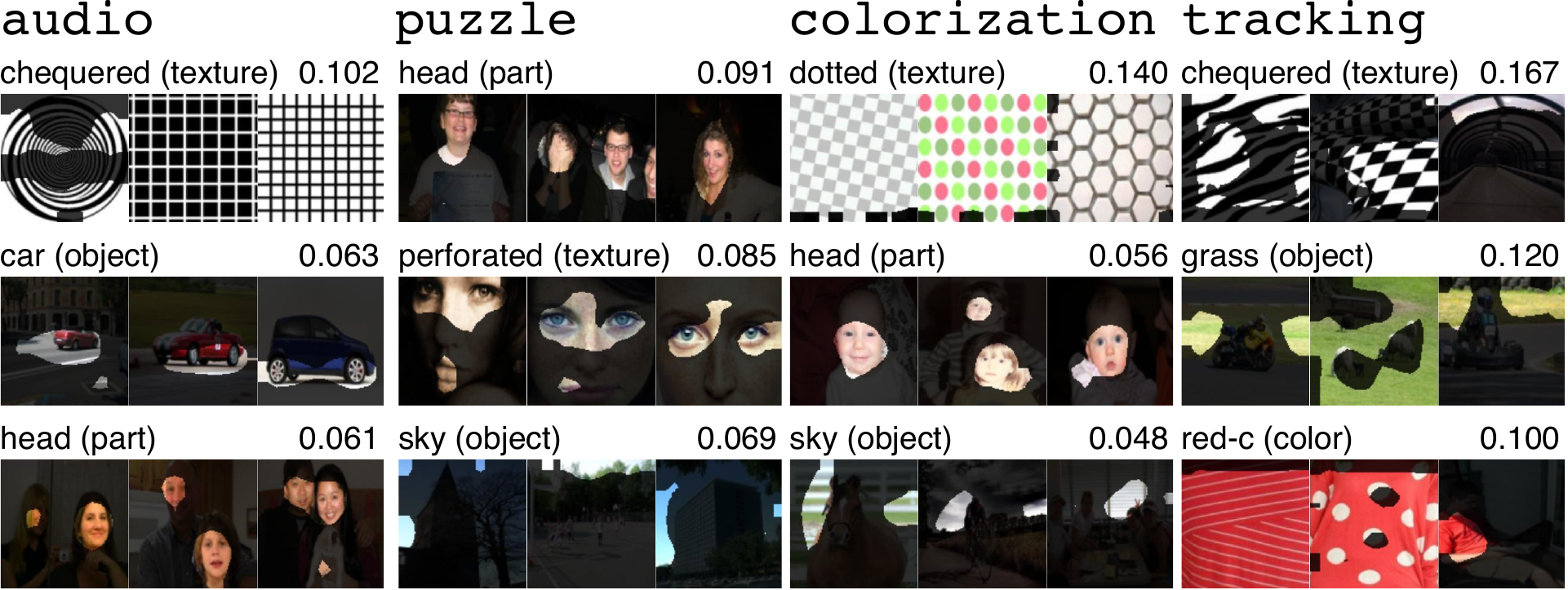}
\vspace{-4mm}
\caption{The top ranked concepts in the three top categories in four self-supervised networks. Some object and part detectors emerge in \texttt{audio}. Detectors for person heads also appear in \texttt{puzzle} and \texttt{colorization}. A variety of texture concepts dominate models with self-supervised training. }\label{fig:selftaught_matter}
\end{figure}

Fig.~\ref{fig:selftaught_matter} shows some typical visual detectors identified in the self-supervised CNN models. For the models \texttt{audio} and \texttt{puzzle}, some object and part detectors emerge. Those detectors may be useful for CNNs to solve the primary tasks: the \texttt{audio} model is trained to associate objects with a sound source, so it may be useful to recognize people and cars; while the \texttt{puzzle} model is trained to align the different parts of objects and scenes in an image. For \texttt{colorization} and \texttt{tracking}, recognizing textures might be good enough for the CNN to solve primary tasks such as colorizing a desaturated natural image; thus it is unsurprising that the texture detectors dominate.

\subsection{Training Conditions vs. Interpretability}

\begin{figure}
\vspace{-1mm}
\begin{center}
\includegraphics[width=1\linewidth,trim={12.6cm 8mm 0 3mm},clip]{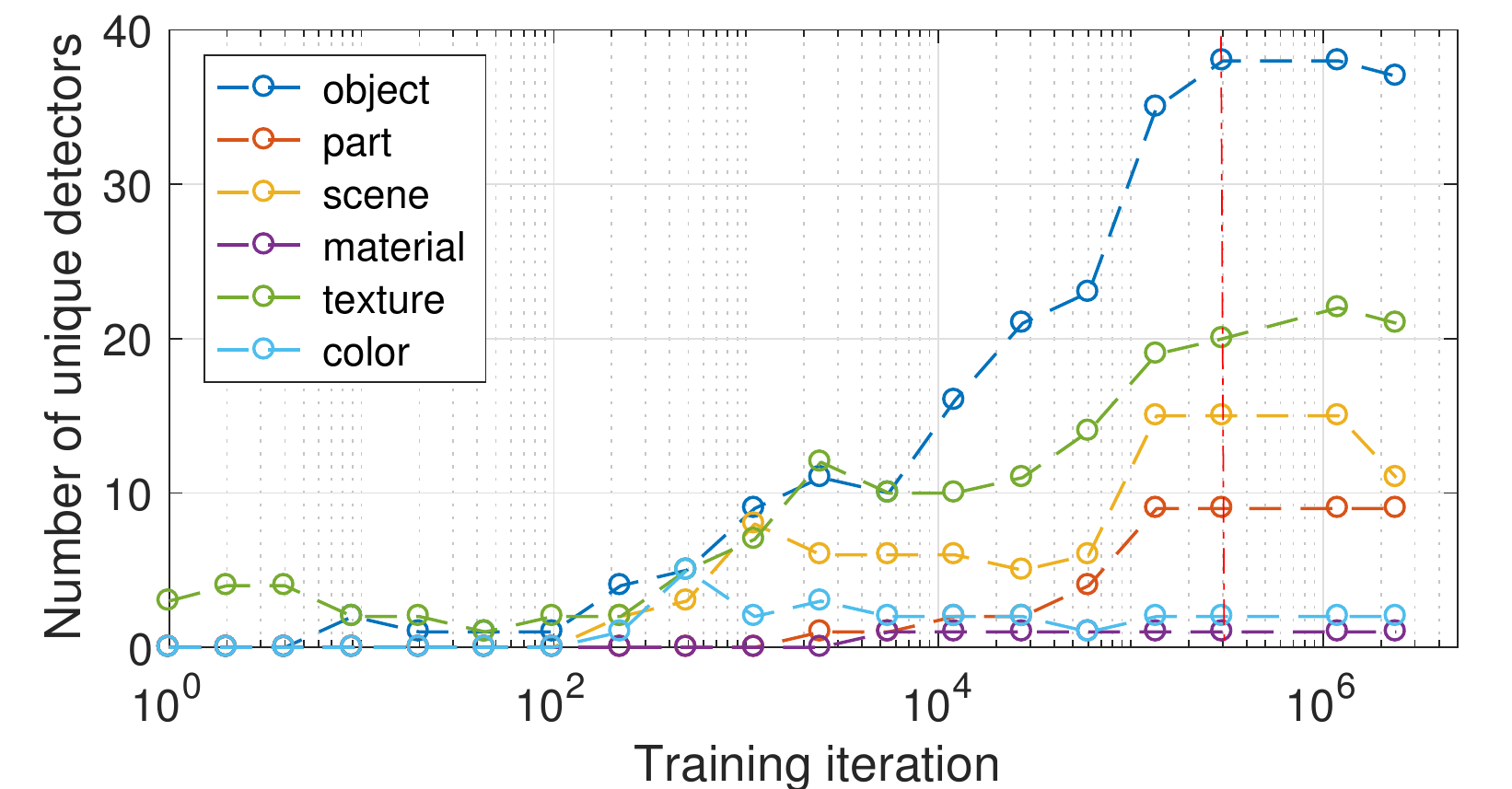}
\end{center}
\vspace{-3mm}
\caption{The evolution of the interpretability of \texttt{conv5} of Places205-AlexNet over 2,400,000 training iterations. The baseline model is trained to 300,000 iterations (marked at the red line).}\label{iterations}
\end{figure}

\begin{figure}
\vspace{-4mm}
\begin{center}
\includegraphics[width=1\linewidth,trim={8mm 4mm 0mm 2mm}]{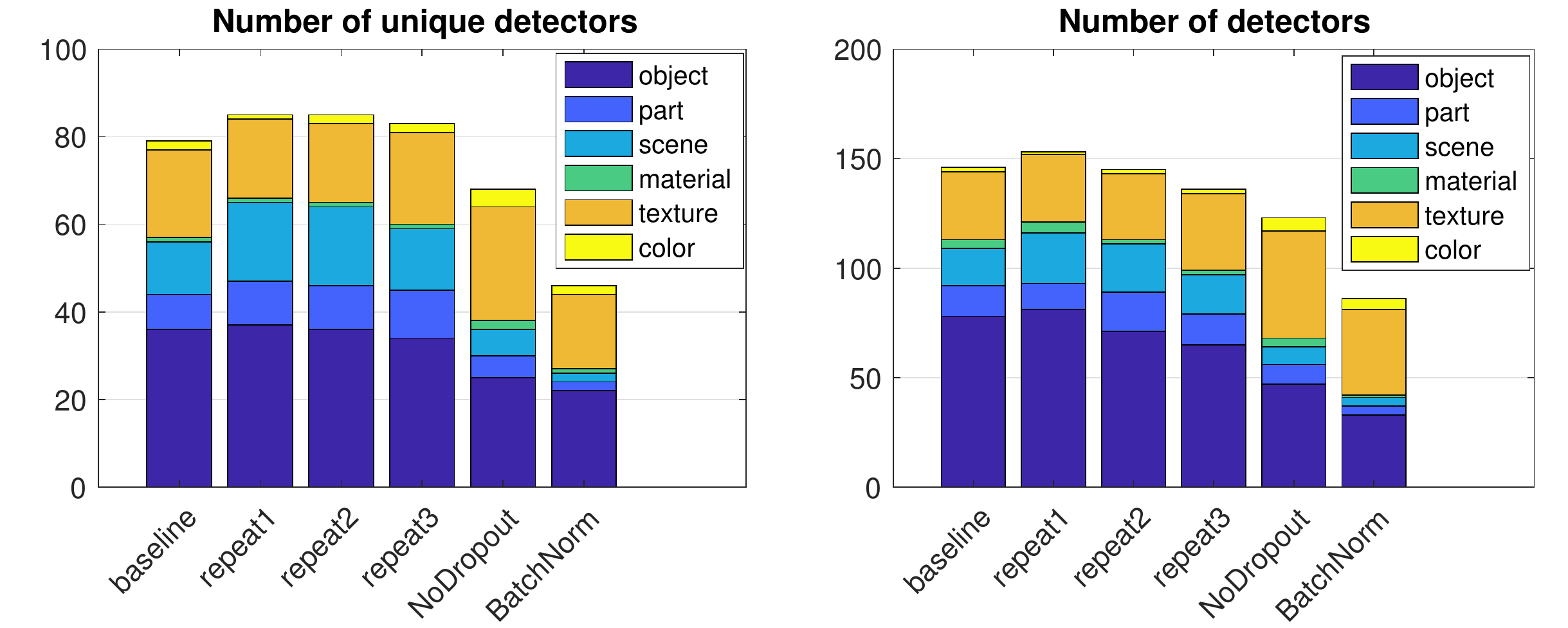}
\end{center}
\vspace{-3mm}
\caption{Effect of regularizations on the interpretability of CNNs. }\label{heuristics}
\vspace{-3mm}
\end{figure}

Training conditions such as the number of training iterations, dropout \cite{srivastava2014dropout}, batch normalization \cite{ioffe2015batch}, and random initialization \cite{li2015convergent}, are known to affect the representation learning of neural networks. To analyze the effect of training conditions on interpretability, we take the Places205-AlexNet as the baseline model and prepare several variants of it, all using the same AlexNet architecture.
For the variants \textit{Repeat1},  \textit{Repeat2} and \textit{Repeat3}, we randomly initialize the weights and train them with the same number of iterations. For the variant \textit{NoDropout}, we remove the dropout in the FC layers of the baseline model. For the variant \textit{BatchNorm}, we apply batch normalization at each convolutional layers of the baseline model. Repeat1, Repeat2, Repeat3 all have nearly the same top-1 accuracy 50.0\% on the validation set. The variant without dropout has top-1 accuracy 49.2\%. The variant with batch norm has top-1 accuracy 50.5\%.

In Fig.~\ref{iterations} we plot the interpretability of snapshots of the baseline model at different training iterations. We can see that object detectors and part detectors begin emerging at about 10,000 iterations (each iteration processes a batch of 256 images). We do not find evidence of transitions across different concept categories during training.  For example, units in \texttt{conv5} do not turn into texture or material detectors before becoming object or part detectors.

Fig.~\ref{heuristics} shows the interpretability of units in the CNNs over different training conditions. We find several effects: 1) Comparing different random initializations, the models converge to similar levels of interpretability, both in terms of the unique detector number and the total detector number; this matches observations of convergent learning discussed in \cite{li2015convergent}. 2) For the network without dropout, more texture detectors emerge but fewer object detectors. 3) Batch normalization seems to decrease interpretability significantly.

The batch normalization result serves as a caution that discriminative power is not the only property of a representation that should be measured.  Our intuition for the loss of interpretability under batch normalization is that the batch normalization `whitens' the activation at each layer, which smooths out scaling issues and allows a network to easily rotate axes of intermediate representations during training.  While whitening apparently speeds training, it may also have an effect similar to random rotations analyzed in Sec.~\ref{rotations} which destroy interpretability. As discussed in Sec.~\ref{rotations}, however, interpretability is neither a prerequisite nor an obstacle to discriminative power.  Finding ways to capture the benefits of batch normalization without destroying interpretability is an important area for future work.

\subsection{Discrimination vs. Interpretability}

Activations from the higher layers of CNNs are often used as generic visual features, showing great discrimination and generalization ability \cite{zhou2014learning,razavian2014cnn}. Here we benchmark deep features from several networks trained on several standard image classification data sets for their discrimination ability on a new task. For each trained model, we extract the representation at the highest convolutional layer, and train a linear SVM with $C=0.001$ on the training data for action40 action recognition task~\cite{yao2011human}. We compute the classification accuracy averaged across classes on the test split.

\begin{figure}
\begin{center}
\includegraphics[width=.95\linewidth]{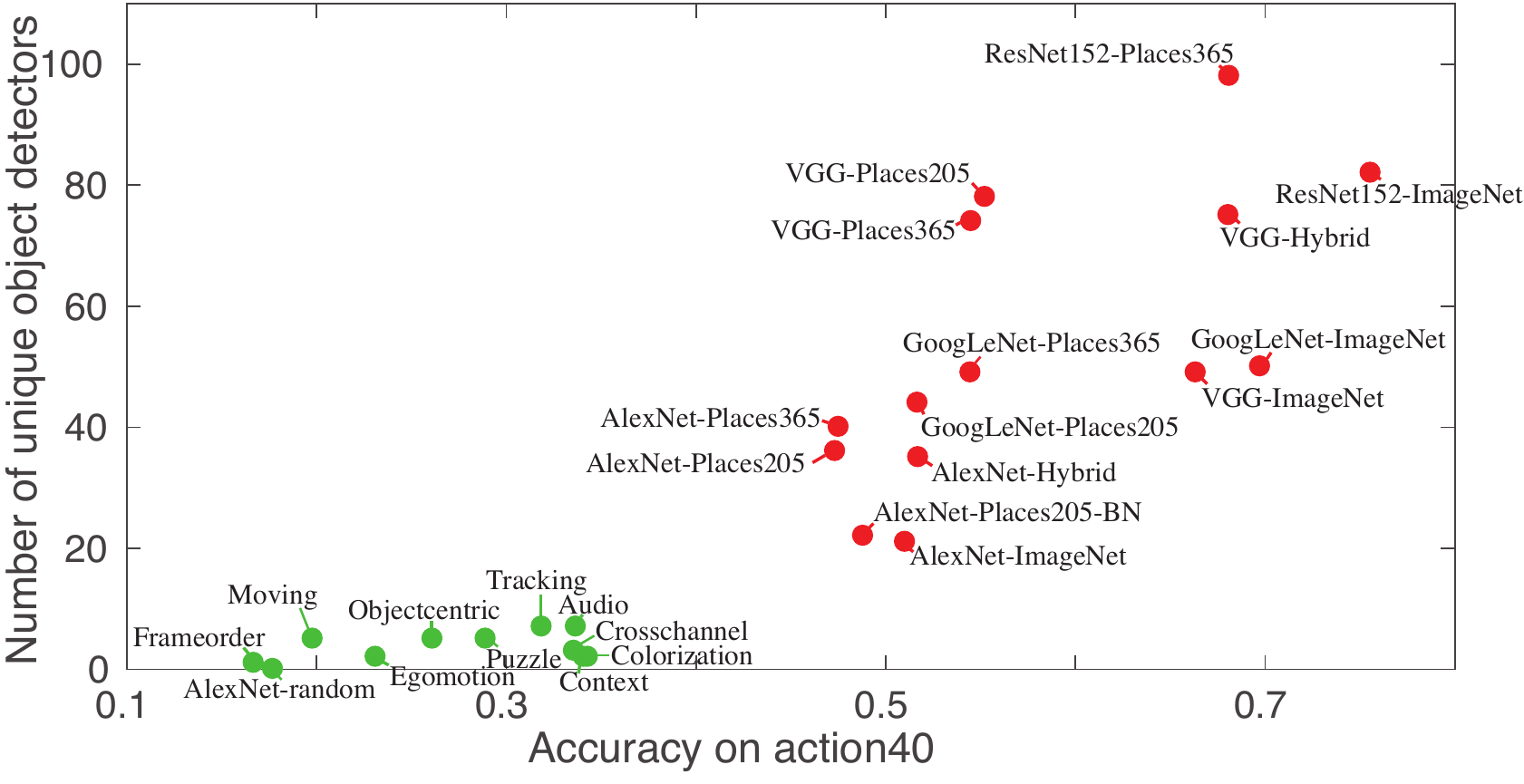}
\end{center}
\vspace{-3mm}
\caption{The number of unique object detectors in the last convolutional layer compared to each representation’s classification accuracy on the action40 data set. Supervised and unsupervised representations clearly form two clusters.}\label{discrimination}
\vspace{-4mm}
\end{figure}

Fig.~\ref{discrimination} plots the number of the unique object detectors for each representation, compared to that representation's classification accuracy on the action40 test set. We can see there is positive correlation between them. Thus the supervision tasks that encourage the emergence of more concept detectors may also improve the discrimination ability of deep features. Interestingly, the best discriminative representation for action40 is the representation from ResNet152-ImageNet, which has fewer  unique object detectors compared to ResNet152-Places365. We hypothesize that the accuracy on a representation when applied to a task is dependent not only on the number of concept detectors in the representation, but on the suitability of the set of represented concepts to the transfer task.

\subsection{Layer Width vs. Interpretability}

From AlexNet to ResNet, CNNs for visual recognition have grown deeper in the quest for higher classification accuracy. Depth has been shown to be important to high discrimination ability, and we have seen in Sec.~\ref{section-comparing-architecture} that interpretability can increase with depth as well. However, the width of layers (the number of units per layer) has been less explored. One reason is that increasing the number of convolutional units at a layer significantly increases computational cost while yielding only marginal improvements in classification accuracy. Nevertheless, some recent work \cite{zagoruyko2016wide} shows that a carefully designed wide residual network can achieve classification accuracy superior to the commonly used thin and deep counterparts.

To explore how the width of layers affects interpretability of CNNs, we do a preliminary experiment to test how width affects emergence of interpretable detectors: we remove the FC layers of the AlexNet, then triple the number of units at the \texttt{conv5}, \textit{i.e.}, from 256 units to 768 units. Finally we put a global average pooling layer after \texttt{conv5} and fully connect the pooled 768-feature activations to the final class prediction. We call this model \textit{AlexNet-GAP-Wide}.

\begin{figure}
\vspace{-3mm}
\includegraphics[width=1\linewidth]{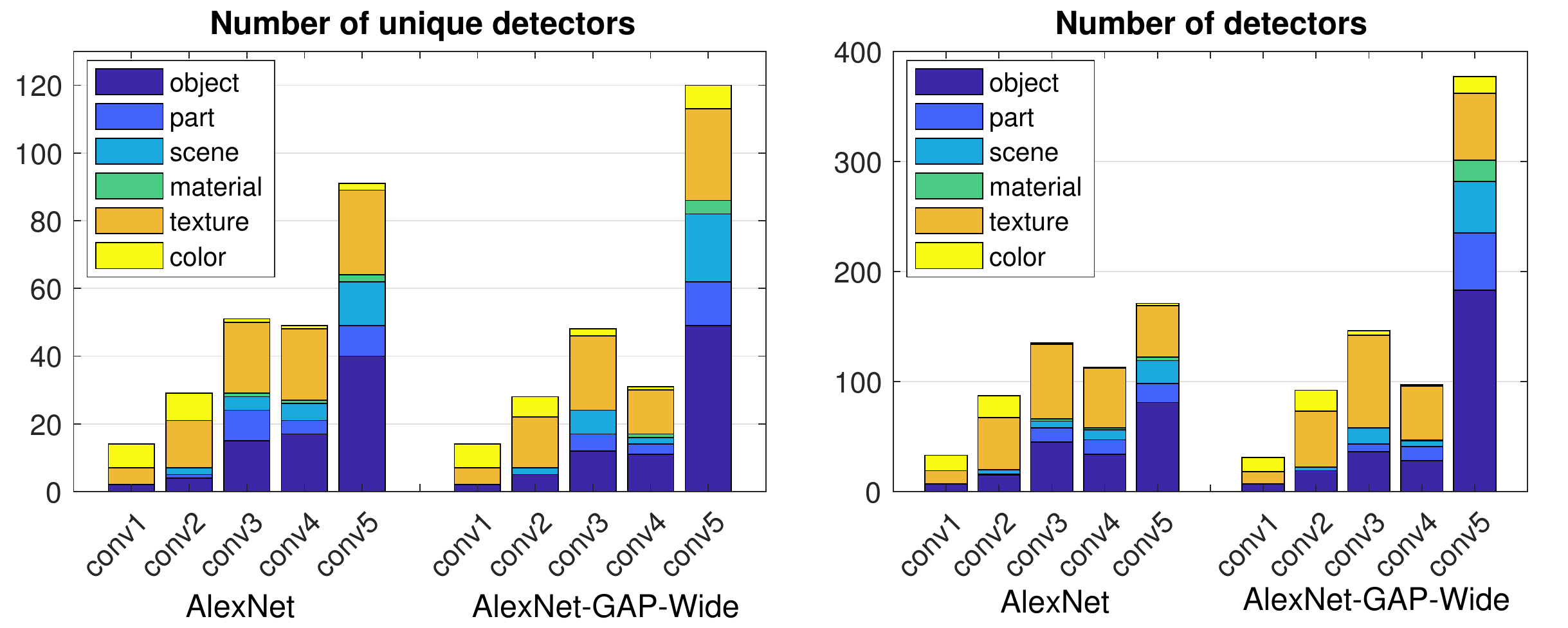}
\vspace{-3mm}
\caption{Comparison between standard AlexNet and AlexNet-GAP-Wide (AlexNet with wider \texttt{conv5} layer and GAP layer) through the number of unique detectors (the left plot) and the number of detectors (the right plot). Widening the layer brings the emergence of more detectors. Networks are trained on Places365.}\label{fig:width_matter}
\vspace{-3mm}
\end{figure}

After training on Places365, the AlexNet-GAP-Wide obtains similar classification accuracy on the validation set as the standard AlexNet ($~0.5\%$ top1 accuracy lower), but it has many more emergent concept detectors, both in terms of the number of unique detectors and the number of detector units at \texttt{conv5}, as shown in Fig.~\ref{fig:width_matter}. We have also increased the number of units to 1024 and 2048 at \texttt{conv5}, but the number of unique concepts does not significantly increase further. This may indicate a limit on the capacity of AlexNet to separate explanatory factors; or it may indicate that a limit on the number of disentangled concepts that are helpful to solve the primary task of scene classification.

\section{Conclusion}

This paper proposed a general framework, network dissection, for quantifying interpretability of CNNs. We applied network dissection to measure whether interpretability is an axis-independent phenomenon, and we found that it is not. This is consistent with the hypothesis that interpretable units indicate a partially disentangled representation. We applied network dissection to investigate the effects on interpretability of state-of-the art CNN training techniques. We have confirmed that representations at different layers disentangle different categories of meaning; and that different training techniques can have a significant effect on the interpretability of the representation learned by hidden units.

\vspace{3mm}
{\footnotesize
\textbf{Acknowledgements}. This work was partly supported by the National Science Foundation under Grant No. 1524817 to A.T.; the Vannevar Bush Faculty Fellowship program sponsored by the Basic Research Office of the Assistant Secretary of Defense for Research and Engineering and funded by the Office of Naval Research through grant
N00014-16-1-3116 to A.O.; the MIT Big Data Initiative at CSAIL, the Toyota Research Institute / MIT CSAIL Joint Research Center, Google and Amazon Awards, and a hardware donation from NVIDIA Corporation. B.Z. is supported by a Facebook Fellowship.
\par}

\normalsize

{\footnotesize
\bibliographystyle{ieee}
\bibliography{mainbib}
}

\end{document}